\documentclass[10pt,a5paper,twoside]{article}
\usepackage{coling2012}
\usepackage{amsmath}
\usepackage{url}
\usepackage{caption}
\usepackage{subcaption}

\title{Detecting English Writing Styles For Non-native Speakers}

\author{$Rami~Al-Rfou'$\\
  Department of Computer Science \\
  Stony Brook University \\
  NY 11794, USA \\
  \texttt{ralrfou@cs.stonybrook.edu}}

\begin{document}
\maketitle
\abstractEn{
  Analyzing writing styles of non-native speakers is a challenging task. In this
  paper, we analyze the comments written in the discussion pages of the English
Wikipedia. Using learning algorithms, we are able to detect
  native speakers' writing style with an accuracy of 74\%. Given the diversity of
the English Wikipedia users and the large number
  of languages they speak, we measure the similarities among their native
languages
  by comparing the influence they have on their English writing style. Our results show that languages known to have the same origin and
  development path have similar footprint on their speakers' English writing style. To
  enable further studies, the dataset we extracted from Wikipedia will be made
  available publicly.
}

\keywordsEn{Stylometric Analysis, Wikipedia, Author attribution, Classification,
Clustering}

\newpage

\section{Introduction}
Stylometric analysis has important applications that cover deception detection,
authorship attribution and vandalism detection \cite{Harpalani, Ott}.
Analyzing English writing styles for non-native speakers is a harder task due to the
influence of their native spoken languages.
Such influence introduces a bias in the orthographical and syntactic errors made by the author and the choice of vocabulary \cite{koppel2005automatically}.

Previous work, in this regard, focused on smaller datasets like {International Corpus of Learner English} (ICLE) \cite{koppel2005automatically, koppel2005determining, argamon2009automatically}. This choice limits the number of the native languages targeted and the set of topics covered.
It also focuses on features that might only appear in the writing styles of
students, e.g. choice of words \cite{tsur2007using, zheng2003authorship,
gamon2004linguistic}. Syntactic features as subject-verb disagreement, mismatch
of noun-number pairs and wrong usage of determiners were studied by
\cite{wong2009contrastive}.
\cite{wong2010parser, wongdras2011EMNLP} used more sophisticated syntactic features as parse trees
to examine the frequency of some distinguishable grammar rules.

Different from previous work, we use different data source. We extract users'
comments from the English Wikipedia talk pages \footnote{Most Wikipedia pages
have corresponding discussion pages which are called talk pages.}. Our dataset is more challenging to study for the following reasons:
First, the comments tend to be shorter in length than the articles from ICLE,
and they cover a diverse spectrum of topics.
Second, Wikipedia users represent a wider
range of fluency in English. Moreover, the
style of the comments is colloquial which limits the choice of features used.
Finally, the targeted languages are more. 
Our contributions in this paper are:
\begin{compactitem}
\item Analyzing common mistakes and patterns of non-native
speakers' writing styles.
\item Studying the similarities among languages using the English writing
styles of the non-native speakers.
\item Publicly available dataset composed of English Wikipedia users' comments.
\end{compactitem}

This paper is structured as follows: Section \ref{wiki} discusses various
aspects of Wikipedia's structure and content. Section \ref{setup} describes
the methodologies used to construct the dataset and filter the noise. In section
\ref{exps_sec}, we discuss the experiments conducted.
Finally, we conclude and present possible avenues of future research.

\section{Wikipedia}
\label{wiki}
Wikipedia is the de facto source of knowledge for internet users. Recently, it
is has been extensively used to help solving different information retrieval
tasks, especially the ones that involve semantic aspects \cite{Milne08aneffective}.
Wikipedia can be used to help
 common NLP tools to perform better; the size of the data and the diversity of
 authors and topics play a key role. Moreover, the sustained growth of Wikipedia
 content can bring performance gains with no additional complexity costs.

With more than 90 thousand active users and 4.4 million articles (in its English
version), Wikipedia spans large number of topics.
Wikipedia pages are saved under a control revision system that keeps track of
users' edits and comments.
Such resource presents a higher quality of data that is not achievable by the other commonly used sources of text as news, blogs and scientific articles.

Wikipedia has a complex database structure to serve its users.
Therefore, extracting data is not trivial. Our goal is to first identify the
language skills of users and then collect their contributions. To identify the
language skills, Wikipedia has an information box called \emph{Babel} that users can
add voluntarily to their profile pages in order to state their skills in different
languages. A user can identify her native language
and her skills in non-native languages on a scale of 0-5.

The task of collecting the contributions of a specific user is a more complex
procedure. The differences among Wikipedia page revisions has to be generated and
linked back to the user table. The resources we have are not sufficient enough to process such
huge amount of data\footnote{Recent efforts were
made to generate the diffs \url{http://dumps.wikimedia.org/other/diffdb/}}.
Instead we noticed that Wikipedia pages have accompanying talk pages where
users discuss different aspects of the articles. In those pages, the style
guideline encourages the user to sign her comments with her own signature that
links back to her user page. The style of writing of these
talk pages are less formal and technical than the main pages of Wikipedia and
has more colloquial features.

\section{Experimental Setup}
\label{setup}
In English Wikipedia, we found that around 60 thousand users specified their language skills, 47\% of
whom are English native speakers. The total number of comments found in the
processed talk pages is around 12 million. Only 2.4 million comments have
users with identified language skills. Since almost half of the users contribute to the
talk pages, the number of users who make at least one comment is around 30 thousand.

Since we have large number of comments and users, we have to filter the user
base to increase the quality of the gathered data. The rules, specified for this
filter, are as follows:
\begin{compactitem}
\item Group the users by their native languages and only consider the
users from the 20 most frequently used languages.
\item English native speakers can pick more fine grained categories, e.g. UK
English, US English, etc. Only
  speakers under the US English category are selected.
\item Users who specified more than one native language are excluded to help
avoid improbable scenarios where users claim to be native in many
languages.
\end{compactitem}

The dataset after filtering the users constitutes of 9857 users and 589228
comments. Comments were filtered according to the following criteria:
\begin{compactitem}
\item Comments need to have at least 20 tokens.
\item Proper nouns are replaced by their Part of Speech (POS) tags to avoid bias toward topics.
\item Non-ASCII characters are replaced by a special character to avoid bias
toward non-English character usage in the comments.
\item The classifier has the same number of comments for each of its classes.
  The two baseline classifiers are: the most common label and the random
  classifier. Each of these will have an accuracy of \verb+1/(number of classes)+.
\item The dataset is split into 70\% training set, 10\% development set and 20\%
as a testing set.
\end{compactitem}

\section{Experiments}
\label{exps_sec}
\subsection{Features}
\label{features}
Given a training dataset, the comments are grouped by classes. The following
n-grams are constructed for each class:
\begin{compactitem}
\item 1-4 grams over the comments' words.
\item 1-4 grams over the comments' characters.
\item 1-4 grams over the part of speech tags of comments' words.
\end{compactitem}
For each class, we will construct $3*4$ n-gram models. For each comment, we will
construct a feature vector of the similarity scores between
the comment and each of the n-gram models. Therefore, if a problem has six
classes, $6*3*4 = 72$ features will be generated for each comment.

For example, the similarity scores ($Sim$) calculated for a comment ($C$)
against the words n-grams models $words\_model(n)$.

\[
  Sim(C, n) = \sum_{x \in grams(C,n)} \log_2 (count(x, n))
\]

\[
  count(x, n) = \left\{
  \begin{array}{l l}
    words\_model(n, word), & x \in words\_model(n)\\
    1, & x \notin words\_model(n)\\
  \end{array} \right.
\]

Other features also include the relative frequency of each of the stop words
mentioned in the comment. The 125 stop words are extracted from the NLTK stop
words corpus\cite{nltk}. Moreover, the average size of words, the size of the
comments and the average number of sentences are also included.

\subsection{Native vs Non-native Experiment}
\label{exps}
This experiment aims to detect the non-native speakers writing styles. The
classifier should be able to distinguish between comments by native speakers
and other non-native speakers.
All users with native language other than English are placed into one category.
The number of comments used is around 322K. Table 
\ref{table:results} shows that the linear SVM classifier reaches
$74.53\%$ accuracy, given the features explained in section \ref{features}.
\begin{table}
  \begin{center}
  \begin{tabular}{l|ll}
	Experiment & Logistic Regression & Linear SVM
	\\\hline
	Non-native & 74.45\% & 74.53\%\\
	Frequent & 50.27\% & 50.26\%\\
	Families & 50.81\% &50.53\% \\
\end{tabular}
\caption{Accuracy of classification using different learning algorithms.}
\label{table:results}
\end{center}
\end{table}

The most informative features are word trigrams, word unigrams, word bigrams,
word 4-grams, character bigrams, POS tag 4-grams, ordered by their importance.
Table \ref{table:nonnative} shows the most correlated grams with native and non-native
speakers.

Analyzing table \ref{table:nonnative}, we can notice that some of the n-grams
indicate common grammatical mistakes in non-native speakers' writing styles.
For example, word unigrams show that non-native speakers tend to use ``earth''
instead of ``Earth''. Character bigrams show that separating the comma from the
previous word by a space is a common usage of punctuation for non-native speakers.
The usage of determiners is a problematic issue for non-native speakers. In the
word 4-grams, we can see a common mistake in the use of ``the'' before a proper
noun. The over-usage of ``the'' can be
validated by looking at the character bigrams where ``th'' appears. Word
trigrams show
that native speakers use ``the'' correctly in ``in the middle'' where we
expect non-native speakers to use ``in middle''.
Moreover, from the character bigrams, non-native speakers use ``at'' less
than the native speakers which might suggest that they incorrectly use other articles
in place of ``at''. Character bigrams show a trend in spelling
mistakes where non-native speakers type single ``l'' instead of ``ll''.
Another mistake is the less frequent usage of apostrophes `` ' '';
this can be traced to more frequent usage of ``am'' in word unigrams for
non-native speakers and the appearance of ``don't'' in the word 4-grams for native speakers.

\begin{table}[t]
  \begin{subtable}[]{0.5\textwidth}
    \begin{tabular}{l|ll|}
     \textbf{Feature} & \textbf{Speaker}  & \textbf{Gram}
     \\\hline
     Words & Non&article on NNP\\
     Trigrams& Native&  \emph{comma} you have\\\cline{2-3}
             & Native& used in the\\
             &  &in the middle\\\hline
       Words &Non& scene, To,\\
     Unigrams&Native&describing, earth,\\\cline{2-3}
             && referenced, am\\\hline
      Words  &Non& You have, we do\\
      Bigrams&Native& of people\\\cline{2-3}
             &Native& years of, I see\\
             &&if you\\\hline
       Words &Non& but I think that\\
     4-grams  &Native&by the NNP of\\\cline{2-3}
              &&( or at least\\
              &Native&NNP on NNP NNP\\
              &&if you don't\\\hline
    Characters&Non& th, e \emph{space},\\
         Bigrams &Native& \emph{space} \emph{comma}\\ \cline{2-3}
                 &Native& l-, at, ll\\\hline
      PoS 4-gram &Non& NNS \textbf{,} DT NN\\
                 &Native& MD VB VBN \textbf{,}\\\cline{2-3}
                 &Native& \textbf{,} PRP VBZ JJ\\
                 && VBP NN IN NN\\
      \end{tabular}
    \caption{Non-native speakers experiment}
    \label{table:nonnative}
  \end{subtable}
  \begin{subtable}[]{0.5\textwidth}
   \begin{tabular}{l|ll}
    \textbf{Feature} & \textbf{Speaker}  & \textbf{Gram}
   \\\hline
   Words & Russian &prove that\\
    Bigrams& German& you leave\\
    & Spanish& to reveal\\
    & Dutch & reliance on\\\hline

   Words &Dutch& refuting\\
      Unigrams&Spanish&timelines\\
       &German& tie\\\hline

   Words&Russian& stick to what\\
   Trigrams&Spanish& find out the\\
   &Dutch& end up in\\\hline

   Characters &French& hee \emph{space}\\
   4-grams      &French&ownr\\
&Dutch&c/es\\
   \hline
   Words &French& NNP as far as\\
   4-grams &German& the NNP who\\
   &Spanish& ,etc),\\
   &EN-US& Does anyone know if\\\hline
   PoS 4-gram &Dutch& RB CD -RRB- IN \\
   &EN-US& VBD , RB DT\\
   &EN-US& CD VBD NNS IN\\
   && \\
   && \\
   && \\
   \end{tabular}
   \caption{Most frequent languages experiment}
   \label{table:frequent}
   \end{subtable}
\label{table:analysis}
\caption{Correlated grams and speakers. For each class and each informative
features the z-scores of each n-gram are calculated. The n-grams with the
highest z-scores are reported in the table.}
\end{table}

\begin{figure}
\centering
\includegraphics[scale=0.50]{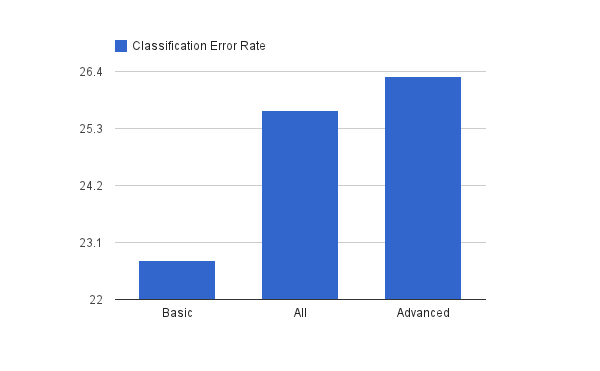}
\caption{Classification error rate against non-native speakers with different
skills.}
\label{fluency}
\end{figure}

The more fluent the speaker is in English, the closer her writing
style to the native speaker style. To test our basic intuition,
we take advantage of the specific language fluency levels that the user
specified in the available Wikipedia Babel box. We designed two different
variations of the previous experiment. In the first, we limited
the non-native speaker's class to the speakers with basic English skills
(identified by the fluency scale 0-2).
Whereas the second variant is composed of the more advanced non-native speakers with
English fluency levels ranging from scale 3-5.
Figure \ref{fluency} shows the classifier error rate in the previous three
variations.

The increase in the error rate of classification confirms our intuition.
Moreover, it increases our confidence in the information given by the users,
regarding their language skills, in their profiles.

\subsection{Frequent Languages Experiment}
\label{frequent}
\begin{figure}

\begin{subfigure}[]{0.5\textwidth}
\includegraphics[scale=0.4]{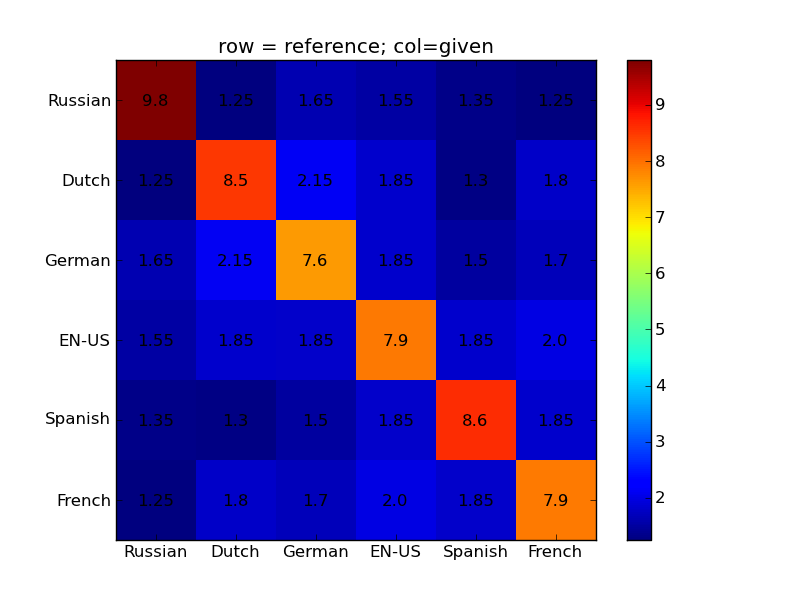}
\caption{Frequent languages experiment}
\label{pop_cfm}
\end{subfigure}
\begin{subfigure}[]{0.5\textwidth}
\includegraphics[scale=0.4]{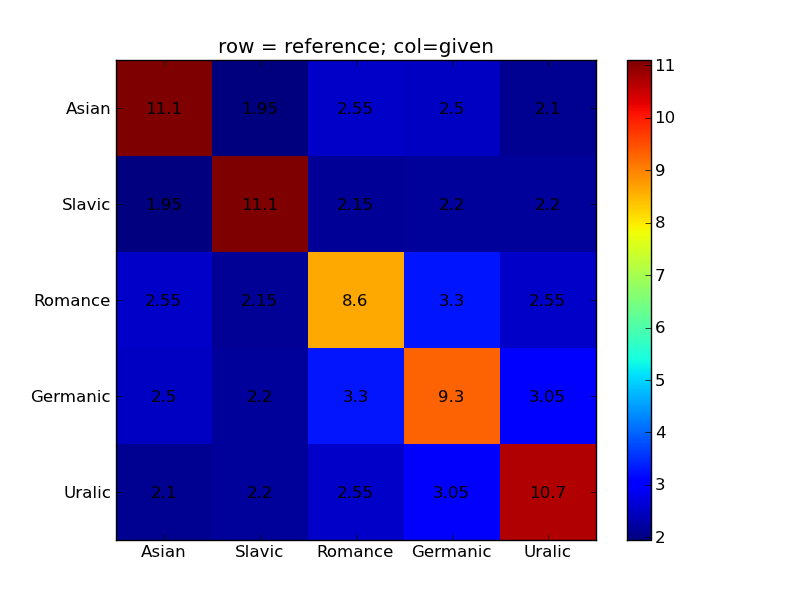}
\caption{Languages families}
\label{fam_cfm}
\end{subfigure}

\caption{Confusion Matrices of different Experiments}
\end{figure}

This experiment aims to classify the comments written by the speakers of the most frequent native languages.
Six languages are selected: US-EN, German, Spanish, French, Russian and Dutch.
Figure \ref{pop_cfm} shows the confusion matrix of the logistic regression
classifier.
Table \ref{table:results} shows that the best accuracy that the classifier
achieved is 50.27\% with 150K comments used. Looking at Figure \ref{pop_cfm}, we can see clearly that the
Russian users are the easiest to identify. Moreover, the classification error is
the highest in distinguishing the German and the Dutch users. These numbers confirm a basic
intuition that the languages that have geographical proximity will have more
borrowed words and grammars among them. Accordingly, this will affect their
speakers' writing styles in English.

The most informative features are ordered as follows: word bigrams, word
unigrams, word trigrams, character 4-grams, word 4-grams, POS tag 4-grams.
We can see that the features of the longer grams become less informative, once we
increased the number of classes given to the classifier because of sparsity.
It may also indicate that there is an influence of the comment's topic on the
classification. Table \ref{frequent} shows some different mistakes made by
different native speakers. For example, for French speakers ``ownr'' was a common
mistake and not adding the space after the comma was another one among the Spanish speakers.

\subsection{Languages Families Experiment}

Looking at experiment \ref{exps}, the confusion in classifying Dutch and German
users suggests that there is a similarity between groups of languages. Referring
to the linguistics research history of classifying the languages into families
according to similar features and development history, this experiment validates
such grouping. The following 18 languages are grouped into 5 families as:
\begin{compactitem}
\item \textbf{Germanic}: German, Dutch, Norwegian, Swedish, Danish.
\item \textbf{Romanace}: Spanish, French, Portuguese, Italian.
\item \textbf{Uralic}: Finnish, Hungarian.
\item \textbf{Asian}: Mandarin, Cantonese, Japanese, Korean.
\item \textbf{Slavic}: Russian, Polish
\end{compactitem}

Figure \ref{fam_cfm} shows that the Slavic and Asian native speakers have a
clear English writing style which is easier to detect. The highest confusion in
classification is between the Germanic and Romance languages, where geographical
proximity plays a role in similarity. With the same reasoning, we can see the confusion between Germanic and Uralic languages.

Taking the opposite approach, we took the speakers of the most frequent 20
native languages and applied the same classification procedure over the new
classes. The accuracy of the classifier is 25\%. However, considering the
confusion matrix as a similarity matrix, we applied the affinity propagation
clustering algorithm \cite{sklearn} over the confusion matrix and the clusters
that were formed are the following:
\begin{compactitem}
\item \textbf{Cluster 1}: Arabic.
\item \textbf{Cluster 2}: Danish, Dutch, Finnish, Norwegian, Swedish.
\item \textbf{Cluster 3}: French, Italian, Portuguese, Spanish.
\item \textbf{Cluster 4}: Mandarin, Cantonese, Japanese, Korean.
\item \textbf{Cluster 5}: Russian, Polish, Turkish.
\item \textbf{Cluster 6}: Hungarian, German, US-EN.
\end{compactitem}

The above clusters, to large extent, support the literature classification of
languages. Scrutinizing the 4-grams POS tags reveal more interesting
observations regarding non-native speaker's usage of English. For example,
in Portuguese speakers' comments, the pattern \verb+IN DT NN PRP+ appears 0.13\%
of the total number of their POS 4-grams. However, it only appears
0.04\% in the Korean speakers' comments. Another observation can be seen by
looking at the \verb+NN NN IN DT+'s usage in Portuguese and Polish comments. It
appears 0.15\% in the former POS 4-grams but only 0.05\% in the later. Arabic
speakers tend to use the pattern \verb+TO DT NN IN+ so frequently that it
appears 0.11\% while it is less than 0.06\% for other speakers' POS 4-grams
distributions. Finally, Japanese and Danish speakers slightly prefer the pattern
\verb+NN PRP VBZ RB+ more than others.

\subsection{Learning Algorithms}

\begin{figure}[t]
\centering
\includegraphics[scale=0.45]{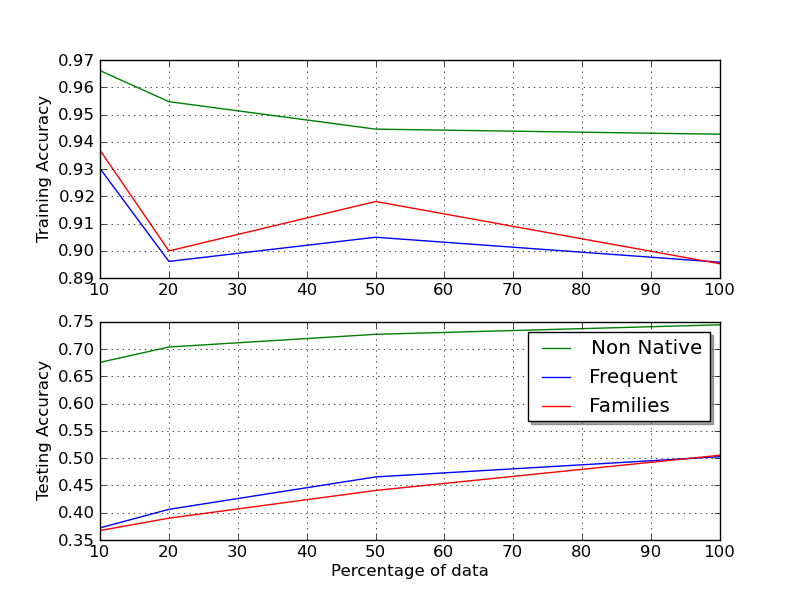}
\caption{Learning curves of the logistic regression algorithm.}
\label{comb_lc}
\end{figure}

Figure \ref{comb_lc} shows a typical over-fitting situation where more data
you have, the better rate the classifier can achieve. Here, the size of data that
can be extracted from Wikipedia plays a significant role to boost the accuracy
from 37\% to over 50\% in case of the Frequent languages and Families of
languages experiments. This confirms the importance of the coverage of words in
the language models that supply the frequency counts on the performance of the classifier.

\section*{Conclusions and Future Work}
\label{conc}

Our results show the effectiveness of the features constructed in detecting writing styles
of a challenging and diverse content. The robustness of the features helped us
in building competitive classifiers on different tasks. Moreover, we were able
to analyze common users' usage patterns of English and discover grammatical and
spelling errors.

Different languages showed different effects on the writing styles of their
speakers. We were able to identify such trends between users' writing styles and
cluster them into groups that supported the well studied origins of languages.

The learning curves show that it is worth increasing the size
of the data in order of magnitude by adding the Wikipedia diffs, especially the
non-minor ones, as it represents another source of users' contributions.

\section*{Acknowledgements}
We would like to thank Steven Skiena for the discussion and the advice.
We are also indebted to the NLTK and the Sklearn teams for producing excellent
NLP and machine learning software libraries.

\newpage
\bibliography{myrefs}{}
\bibliographystyle{apa}

\end{document}